\documentclass{esannV2}
\usepackage{graphicx}
\usepackage[latin1]{inputenc}
\usepackage{amssymb,amsmath,array}

\usepackage[font = footnotesize, labelfont = bf]{caption}
\usepackage[font=footnotesize]{subcaption}
\usepackage{gensymb}        

\usepackage{xcolor}
\usepackage{tikz}
\usetikzlibrary{positioning, fit, shapes.geometric,arrows,calc}

\begin{document}
\title{A Relational Model for One-Shot Classification}

\author{Arturs Polis and Alexander Ilin
%
%
\vspace{.3cm}\\
%
Aalto University, Espoo, Finland
%
}

\maketitle

\begin{abstract}
We show that a deep learning model with built-in relational inductive bias can bring benefits to sample-efficient learning, without relying on extensive data augmentation.
The proposed one-shot classification model performs relational matching of a pair of inputs in the form of local and pairwise attention.
Our approach solves \emph{perfectly} the one-shot image classification Omniglot challenge. Our model exceeds human level accuracy, as well as the previous state of the art, with no data augmentation.


\end{abstract}

\section{Introduction}






Humans can often learn to distinguish previously unseen visual objects after being shown only a few example images of a novel object. Deep neural networks, on the other hand, typically require thousands of training images to generalize to new examples of classes seen during training.
The field of \textit{few-shot learning} aims to address this by developing models that can work with scarce data: only a few examples being available for training. One-shot classification is a special case of few-shot learning, where the system needs to learn to classify after seeing only one example of each class. Although deep learning models have made great progress in few-shot learning~\cite{vinyals2016matching, finn2017maml, snell2017protonets}, reaching human level performance has been hard~\cite{lake2019omniglot} even on simple datasets like Omniglot~\cite{lake2015human} which is one the main benchmarks for this task.


In this work, we propose to use a deep learning model with relational inductive bias~\cite{battaglia2018relational} for the task of one-shot classification.
The proposed model uses self-attention and cross-attention blocks of the Transformer~\cite{vaswani2017transformers} to compare a pair of images. Relational models with Transformer-based attention have recently been shown to succeed in vision tasks \cite{dosovitskiy2020image,caron2021emerging,sarlin2019superglue}
and in this paper we show that this type of models can perform well in few-shot learning. Our proposed model is to the best of our knowledge the first deep learning model that achieves perfect (and better than human) accuracy in the most challenging Omniglot classification setup: one-shot within alphabet classification with 20 alternative classes without using data augmentation.
\begin{figure*}[t]
\begin{minipage}[c]{0.33\textwidth}
    \centering
    \includegraphics[width=0.8\linewidth]{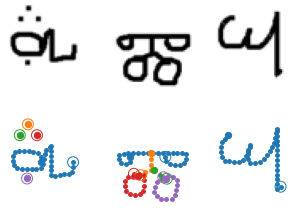}
    \caption{Omniglot data as images and points. Points that belong to the same pen stroke share color. Larger points mark the beginning, and empty circles -- the end of the motion of the pen.}
    \label{fig:points}
\end{minipage}
\hfill
\begin{minipage}[c]{0.65\textwidth}
    \centering
    \captionof{table}{Omniglot within alphabet one-shot, 20-way classification accuracy on the Evaluation set. Full and Min. denote the type of Background set used for training. We show the previously published results, when available, for: Human and BPL~\cite{lake2015human, lake2019omniglot}, Prototypical Networks~\cite{snell2017protonets, lake2019omniglot} and ARC~\cite{shyam2017arc}.}
    \footnotesize
    \begin{tabular}[b]{l l l l l} 
        \hline
        &  \multicolumn{2}{c}{Unaugmented} & \multicolumn{2}{c}{Augmented} \\
        Model & Full& Min. & Full& Min.\\  
        \hline
        Human & $95.5$ & -- & -- & --  \\ 
        BPL & $96.7$ & $\mathbf{95.8}$ & -- & -- \\
        \hline
        Prototypical Networks &  $86.3$ & $69.9$ &$94$ & --\\
        ARC & -- & -- & $97.75$ & --\\
        ARC (full context) & -- & -- & $98.5$ & --\\
        \hline
        AM (ours) -- points &$85.5$ & $44.7$   &$94$ & $74.1$\\
        AM (ours) -- images  & $97.5$& $79$ & $98.25$ & $95.1$ \\
         \begin{tabular}{@{}c@{}}AM (ours) -- images\\  (full context)\end{tabular} & $\mathbf{100}$& $81.38$ & $\mathbf{99.5}$ & $\mathbf{98.5}$ \\
        \hline
    \end{tabular}
    \label{table:omniglot}
\end{minipage}
\end{figure*}



\section{Omniglot challenge}

Lake et al.~\cite{lake2015human} proposed Omniglot benchmark to measure progress of generalization from few examples in various tasks such as classification, generation of new examples and parsing objects into parts. Omniglot dataset contains characters from 50 alphabets, 1623 character classes in total, each character represented by only 20 hand-drawn instances, available both in image and pen trajectory formats  
(a few samples are shown in Fig.~\ref{fig:points}).
The dataset is split into a 30-alphabet Background set (with 964 classes) used for training, and a 20-alphabet Evaluation set (with 659 classes). Recently, the ``Minimal''~\cite{lake2019omniglot} version of the training data has been proposed, containing two five-alphabet subsets of the Background set.
The evaluation procedure of the $N$-way one-shot classification task consists of multiple episodes. Each episode is a classification task with $N$ classes from the same alphabet. The classifier can use $N$ \textit{support} examples (one example per class) to assign $N$ \textit{query} examples to one of the $N$ classes. The Background set can be used for training. For evaluation, we use pre-defined episodes from the Evaluation set with $N=20$, which were proposed by the authors of Omniglot~\cite{lake2015human}.

The authors of Omniglot developed a generative probabilistic model called Bayesian Program Learning (BPL)\footnote{https://github.com/brendenlake/BPL}, as a non-neural baseline for Omniglot tasks.
BPL has held the state of the art for the most challenging setup: the 20-way one-shot classification task without the use of data augmentations during training, it also generalizes very well from the Minimal training set
(see Table~\ref{table:omniglot}).
The authors argue \cite{lake2019omniglot} that the challenge remains unsolved because the human level on the \emph{within-alphabet} classification has been out of reach for deep learning models that do not rely on heavy data augmentation.
Most of the published research on Omniglot classification addresses a simpler \emph{between-alphabets} classification task, in which testing episodes contain characters from different alphabets, making the examples easier to distinguish.

\begin{figure*}
\usetikzlibrary{backgrounds}
\begin{minipage}{\textwidth}
    \begin{minipage}[b]{0.4\textwidth}
        \centering
        \definecolor{color1}{HTML}{f7cac9}
        \definecolor{color2}{HTML}{92a8d1}
        \definecolor{color3}{HTML}{f4e1d2}
        \definecolor{color4}{HTML}{ffef96}
        \definecolor{color5}{HTML}{d5f4e6}
        \definecolor{color6}{HTML}{b9b0b0}
        \definecolor{color7}{HTML}{c2d4dd}
        \definecolor{color8}{HTML}{b8a9c9}
        \definecolor{color9}{HTML}{e06377}

        \tikzset{%
          block/.style    = {draw, thick, rectangle, minimum height = 3em, minimum width = 3em},
          neuron/.style   = {draw, circle, minimum height=2em},
          sum/.style      = {draw, circle, node distance = 1cm}, 
          pixel/.style    = {draw, rectangle, thick, minimum height=1em, minimum width=1em, anchor=center},
          mlayer/.style   = {rectangle, draw=black,
                             minimum height=.5em, minimum width = 2em, text centered},
        }
        
        \begin{tikzpicture}[thick, node distance=7mm]
        
        \draw
        node at (0,0) [rectangle, draw, minimum height=10mm, minimum width=10mm] (i1){\includegraphics[width=0.8cm]{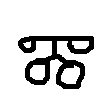}}
        node [trapezium, below of=i1, node distance=10mm, trapezium angle=-60, minimum width=10mm, draw, thick] (cnn1) {\tiny CNN}
        
        node at ($(i1.south) + (0, -10mm)$)[pixel, fill=color1, anchor=north east] (x11) {}
        node [pixel, fill=color2, right of=x11, node distance=1em] (x12) {}
        node [pixel, fill=color3, below of=x11, node distance=1em] (x21) {}
        node [pixel, fill=color4, right of=x21, node distance=1em] (x22) {}
        
        node at ($(x21.south east) + (0, -4mm)$)[pixel, fill=color2, anchor=north east] (x2) {}
        node [pixel, left of=x2, fill=color1, node distance=1em] (x1) {}
        node [pixel, fill=color3, right of=x2, node distance=1em] (x3) {}
        node [pixel, fill=color4, right of=x3, node distance=1em] (x4) {}
        
        node at ($(x21.south east)!0.5!(x2.north east)$){\scriptsize flatten + pos.\ enc.}
        
        node at ($(x2.south east) + (0, -7mm)$)[mlayer, minimum width=18mm] (sa1) {\tiny self-attention}
        node [mlayer, below of=sa1, minimum width=18mm] (ca1) {\tiny cross-attention}

        node [rectangle, right of=i1, node distance=25mm, draw, minimum height=10mm, minimum width=10mm] (i2){\includegraphics[width=0.8cm]{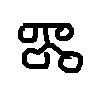}}
        node [trapezium, below of=i2, node distance=10mm, trapezium angle=-60, minimum width=10mm, draw, thick] (cnn2) {\tiny CNN}
        
        node at ($(i2.south) + (0, -10mm)$)[pixel, fill=color5, anchor=north east] (y11) {}
        node [pixel, fill=color6, right of=y11, node distance=1em] (y12) {}
        node [pixel, fill=color7, below of=y11, node distance=1em] (y21) {}
        node [pixel, fill=color8, right of=y21, node distance=1em] (y22) {}
        
        node at ($(y21.south east) + (0, -4mm)$)[pixel, fill=color6, anchor=north east] (y2) {}
        node [pixel, left of=y2, fill=color5, node distance=1em] (y1) {}
        node [pixel, fill=color7, right of=y2, node distance=1em] (y3) {}
        node [pixel, fill=color8, right of=y3, node distance=1em] (y4) {}
        
        node at ($(y21.south east)!0.5!(y2.north east)$){\scriptsize flatten + pos.\ enc.}
        
        node at ($(y2.south east) + (0, -7mm)$)[mlayer, minimum width=18mm] (sa2) {\tiny self-attention}
        node [mlayer, below of=sa2, minimum width=18mm] (ca2) {\tiny cross-attention}
        
        node at ($(ca1.south)!0.5!(ca2) + (0, -9mm)$)[mlayer, minimum width=45mm] (g) {\scriptsize aggregate}
        
        node [below of=g] (score) {\scriptsize score}
        ;
        
        \draw [draw=black] ($(sa1.north west) + (-3mm, 3mm)$)
             rectangle ($(ca2.south east) + (3mm, -3mm)$);
             
        \node at ($(sa1.north) + (1.25cm, 0.1)$)[]{\scriptsize $2 \times$};
        
        \draw [->,>=stealth] (i1.south) -- (cnn1.north);
        \draw [->,>=stealth] (cnn1.south) -- (x11.north east);
        
        \draw [->,>=stealth] (x1.south) -- (sa1.north-|x1.south);
        \draw [->,>=stealth] (x2.south) -- (sa1.north-|x2.south);
        \draw [->,>=stealth] (x3.south) -- (sa1.north-|x3.south);
        \draw [->,>=stealth] (x4.south) -- (sa1.north-|x4.south);
        
        \draw [->,>=stealth,very thick] (sa1.south) -- (ca1.north);
        \draw [->,>=stealth,very thick] (sa1.south) -- (ca2.north west);
        
        \draw [->,>=stealth] (ca1.south-|x1.south) -- (g.north-|x1.south);
        \draw [->,>=stealth] (ca1.south-|x2.south) -- (g.north-|x2.south);
        \draw [->,>=stealth] (ca1.south-|x3.south) -- (g.north-|x3.south);
        \draw [->,>=stealth] (ca1.south-|x4.south) -- (g.north-|x4.south);

        
        \draw [->,>=stealth] (i2.south) -- (cnn2.north);
        \draw [->,>=stealth] (cnn2.south) -- (y11.north east);
        
        \draw [->,>=stealth] (y1.south) -- (sa2.north-|y1.south);
        \draw [->,>=stealth] (y2.south) -- (sa2.north-|y2.south);
        \draw [->,>=stealth] (y3.south) -- (sa2.north-|y3.south);
        \draw [->,>=stealth] (y4.south) -- (sa2.north-|y4.south);
        
        \draw [->,>=stealth,very thick] (sa2.south) -- (ca2.north);
        \draw [->,>=stealth,very thick] (sa2.south) -- (ca1.north east);
        
        \draw [->,>=stealth] (ca2.south-|y1.south) -- (g.north-|y1.south);
        \draw [->,>=stealth] (ca2.south-|y2.south) -- (g.north-|y2.south);
        \draw [->,>=stealth] (ca2.south-|y3.south) -- (g.north-|y3.south);
        \draw [->,>=stealth] (ca2.south-|y4.south) -- (g.north-|y4.south);
        
        \draw [->,>=stealth] (g.south) -- (score);
        
        \end{tikzpicture}
        \\
        (a)
    \end{minipage}
    \hfill
    \begin{minipage}[b]{0.20\textwidth}
        \definecolor{grey}{HTML}{e5e5e5}
        \definecolor{pink}{HTML}{fce1e0}
        \definecolor{yellow}{HTML}{fce9a7}
        \definecolor{blue}{HTML}{c1e4f7}
        \definecolor{green}{HTML}{c7e8ac}
        \definecolor{mauve}{HTML}{d1bcd2}

        \centering
        \tikzset{%
          encoder/.style  = {draw, thick, rectangle, rounded corners=1ex},
          tblock/.style   = {draw, thick, rectangle, minimum width = 18mm, rounded corners=0.5ex},
          residual/.style = {draw, thick, rectangle, minimum width = 0.5cm, rounded corners=2mm},
          arrow/.style = {draw, -latex', thick},
          line/.style = {draw, thick},
          support/.style = {coordinate,join=by fuzzy},
        }
        
        \tikzset{fit margins/.style={/tikz/afit/.cd,#1,
                /tikz/.cd,
                inner xsep=\pgfkeysvalueof{/tikz/afit/left}+\pgfkeysvalueof{/tikz/afit/right},
                inner ysep=\pgfkeysvalueof{/tikz/afit/top}+\pgfkeysvalueof{/tikz/afit/bottom},
                xshift=-\pgfkeysvalueof{/tikz/afit/left}+\pgfkeysvalueof{/tikz/afit/right},
                yshift=-\pgfkeysvalueof{/tikz/afit/bottom}+\pgfkeysvalueof{/tikz/afit/top}},
                afit/.cd,left/.initial=2pt,right/.initial=2pt,bottom/.initial=2pt,top/.initial=2pt}
        
        \begin{tikzpicture}[node distance=5mm]
        \node (out) [support] {};
        \node (res) [below=of out, circle, draw, thick, inner sep=0pt, fill=white] {\bf +};
        \node (norm) [below=of res, tblock, fill=yellow] {\scriptsize Norm};
        \node (mlp) [below=of norm, tblock, fill=blue] {\scriptsize  MLP};
        \node (cat) [below=of mlp, residual, fill=white] {\scriptsize  Concat};
        \node (attn) [below=of cat, tblock, fill=green] {\scriptsize\shortstack{Multi-Head\\Attention}};
        
        \begin{scope}[on background layer]
           \node (enc) [encoder, fit margins={left=2.5pt,right=5.5pt,bottom=4.5pt,top=2pt}, fit={(res) (norm) (mlp) (cat) (attn)}, fill=grey] {};
        \end{scope}
        
        \node (after feats) [below of=attn, support, node distance=5mm] {};
        \node (feats) [below of=after feats, tblock, fill=pink, yshift=-2mm] {\scriptsize  Features};
        
        \path [line] (feats.north) -- (attn.south);
        \path [arrow, rounded corners=4pt] (after feats) -- +(1.1, 0)  |- (cat.east);
        \path [arrow, rounded corners=4pt] (after feats) -- +(1.1, 0)  |- (res.east);
        \path [arrow] (attn.north) -- (cat.south);
        \path [arrow] (cat.north) -- (mlp.south);
        \path [arrow] (mlp.north) -- (norm.south);
        \path [arrow] (norm.north) -- (res.south);
        \path [arrow] (res.north) -- (out.south);

        \end{tikzpicture}
        \\[5mm]
        (b)
    \end{minipage}
    \hfill
    \begin{minipage}[b]{0.30\textwidth}
        \centering\small
        \begin{tabular}{c|c}
        from $A$:& to $B$: \\
        \hline
        \includegraphics[trim=0 0 0 -5, width=0.4\linewidth]{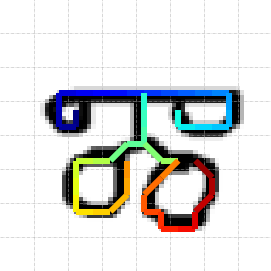} &
        \includegraphics[trim=0 0 0 -5, width=0.4\linewidth]{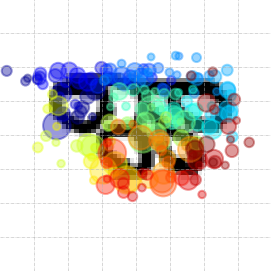}\\
        
        \includegraphics[trim=0 0 0 -5, width=0.4\linewidth]{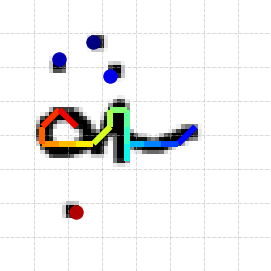} &
        \includegraphics[trim=0 0 0 -5, width=0.4\linewidth]{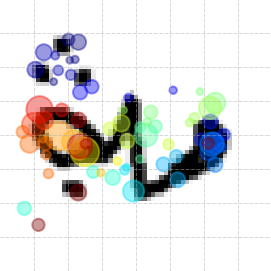}\\

        \includegraphics[trim=0 0 0 -5, width=0.4\linewidth]{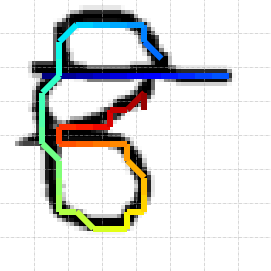} &
        \includegraphics[trim=0 0 0 -5, width=0.4\linewidth]{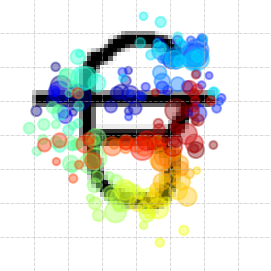}\\
        
        \includegraphics[trim=0 0 0 -5, width=0.4\linewidth]{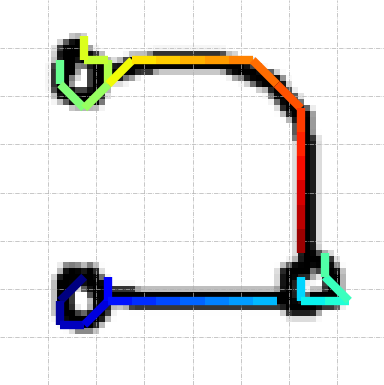} &
        \includegraphics[trim=0 0 0 -5, width=0.4\linewidth]{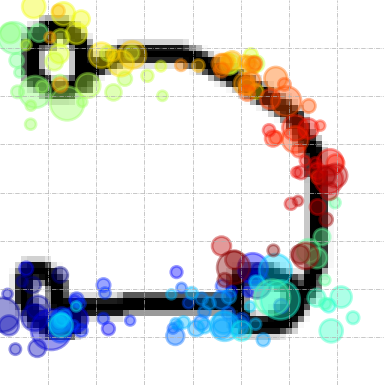}\\
        
        \end{tabular}
        \\[3mm]
        (c)
    \end{minipage}
\end{minipage}
\caption{(a): The architecture of the proposed AttentiveMatcher model.
(b): The attention layer used in our model.
(c): Cross-attention strength from image $A$ to image $B$, colors show which regions in $A$ attend to which regions in $B$. The corresponding parts match despite being drawn quite differently.}
\label{fig:architecture}
\end{figure*}

\section{AttentiveMatcher: the proposed model}
As humans, when we compare two unfamiliar images to see if they show the same item, we first look at each image to verify its contents, and then compare parts of each image to one other to see if all parts match. The same idea informs the approach presented in this paper: our model uses self-attention to learn the structure of each image, and cross-attention to match the corresponding parts between the two images, then the model aggregates this information to determine if images belong to the same class. With the above intuition in mind we call our model AttentiveMatcher (AM).

The training of AttentiveMatcher is performed on pairs of input, $A$ and $B$,
to predict whether they belong to the same class or not.
The architecture of the model is illustrated in Fig.~\ref{fig:architecture}a. 
We use a convolutional network to embed the input images, obtaining a grid of $16 \times 16$ super-pixels, which we treat as $256$ features. Similar to the hybrid model in~\cite{dosovitskiy2020image} we use these features as inputs to the Transformer layers.
First, we separately perform self-attention on the features coming from the two images. Next, we perform cross-attention between the features in the two inputs. We apply the above sequence of attention operations twice, in order for information in each feature to propagate across both inputs. 

As the attention blocks, we use the Transformer~\cite{vaswani2017transformers} attention layers with the modifications inspired by~\cite{sarlin2019superglue}. Unlike in Transformer, where the outputs of the multi-head attention blocks are passed through a feed-forward block directly, we first concatenate the attention outputs with the input features before passing them to the feed-forward block followed by Layernorm~\cite{ba2016layer} (see Fig.~\ref{fig:architecture}b). The feed-forward multilayer perceptron (MLP) is then followed by a residual connection.
All the MLPs in our network have one ReLU-activated hidden layer.

After performing two rounds of self and cross-attention, we aggregate all the transformed local features into a single scalar score predicting the match between $A$ and $B$. The aggregation is inspired by the Graph Matching Networks~\cite{li2019gmn}:
\begin{equation}
    s = h \left( \sum_{i} \ 
    \boldsymbol{\alpha}_i \odot \
    f (\mathbf{z}_i) \right),
    \quad
    (\boldsymbol{\alpha}_i)_k = \frac{\exp g_k(\mathbf{z}_i)}{\sum_j \exp g_k(\mathbf{z}_j)}
\,,
\label{eq:emb}
\end{equation}
where $\mathbf{z}_i$ are the local features from the two inputs, $(\boldsymbol{\alpha}_i)_k$ is the $k$-th element of $K$-dimensional vector $\boldsymbol{\alpha}_i$ and functions $f, g_k, h$ are modeled with MLPs.
The output layer of $h$ has the sigmoid function, which results in score $s \in (0, 1)$ treated as the probability that the two inputs belong to the same class. The training batches are constructed by sampling positive (same class) and negative (different classes) pairs from the training set. We use the binary cross-entropy loss to tune the parameters of the model. 

The testing episodes are constructed by pairing each query input with each support input, resulting in $N\times N$ comparison pairs for each episode. 
Each query example is assigned a label of the support example with the highest score $s$.
The model accuracy is defined as the fraction of the labels correctly assigned.
To compare our model with ARC~\cite{shyam2017arc}, which is the previous state of the art for the task, we also experimented in the scenario assumed by ARC when the query examples in one episode are known to belong to different classes. We used Hungarian matching~\cite{kuhn1955hungarian} which assigns each support example to a unique query example in the episode, while maximizing overall score.

\section{Experiments and results}
\label{sec:results}

We perform one-shot classification
for 20-way Omniglot within alphabet task.
Images are resized to $64 \times 64$ and passed through a four-layer convolutional network commonly used in few-shot models \cite{snell2017protonets, sung2018compare}. 
We use kernel of size $3$, with padding and the stride of $1$, and we use max-pooling only after the first two convolutional layers.
This results in a $16 \times 16$ feature map with $d_f=128$ channels which we flatten into a sequence of $256$ local features of size $d_f$ that we use as input tokens to the Transformer layers. We use simple 1D positional encoding~\cite{vaswani2017transformers} for the inputs to the Transformer block and we use $K=2048$ dimensions in the aggregation network in \eqref{eq:emb}.

We run our experiments on both unaugmented and augmented data. When augmenting, we add rotations at 90{\degree}, 180{\degree}, 270{\degree} as extra training classes, similar to~\cite{snell2017protonets}, resulting in 3856 classes.
%
We use the batch size of 256 and we construct mini-batches such that the contribution of the positive examples in the loss is similar to the contribution of the negative examples. We use the Adam~\cite{kingma2014adam} optimizer with initial learning rate of $0.0001$ reduced by factor of $0.1$ every $100$ epochs, and clip gradients at $2.5$. While training, we validate by forming random within-alphabet episodes from the ten alphabets not in the test set. We stop the training once validation accuracy stops improving.

Additionally, we test our model with input characters represented as sequences of points. To extract points from Omniglot strokes, we follow the pre-processing done by BPL and linearly interpolate for equal spacing with $5$ pixels between neighbour points.
We represent inputs as sequences of point coordinates. 
Each point is embedded using an MLP to size $d_f=64$ and followed by dropout. We use $K=1024$ in the aggregation network in \eqref{eq:emb}.



Table~\ref{table:omniglot} contains the obtained results.
Our unaugmented model outperformed other published results, including BPL~\cite{lake2015human} and Prototypical Networks~\cite{snell2017protonets} and achieved perfect accuracy after applying Hungarian algorithm at test time, denoted by ``full context'' in the results. When training on the Minimal set, our model benefits from augmentation, and when full context is used, outperforms the previous state of the art, BPL. The model trained on points using data augmentation (three extra rotations per character) shows comparable accuracy to our image model, while the unaugmented results on points fall short of the corresponding image results. This is somewhat surprising, as one would assume that stroke trajectories contain the same amount of relevant information as the images. A likely reason for images performing better is the translation invariance induced by the convolutional network helping model to generalize better when compared to point-features prepared by simple MLPs. 



The model allows us to visualize the attention, as shown in Fig.~\ref{fig:architecture}c. The cross-attention patterns obtained on Omniglot are meaningful: the classifier matches parts of its two inputs that represent the same strokes. 

We also tested our model on natural images using the popular miniImageNet~\cite{vinyals2016matching} benchmark.
The model works well but does not exceed the state of the art. The performance of our model on natural images could also have been constrained by the small model size. We also studied the cross-attention patterns on the miniImageNet data and found that
the model paid more attention to similarities in textures rather than semantically similar high-level parts and objects.

\section{Conclusion}



Our model obtains a perfect accuracy on the challenging Omniglot within alphabet classification task, exceeding both human level and BPL on unaugmented data, and to the best of our knowledge is the first model to do so. 
Our work shows that the relational inductive bias, in the form of local and pairwise attention, can be useful in solving few-shot learning problems.
More work is needed to investigate the potential of this approach on natural images, where it is more challenging to develop meaningful relational representations in terms of objects and object parts.
In addition, the relation between model size and interpretability of attention patterns for input matching should be explored. 
Working in this direction will drive the improvement in deep learning methods for efficient and scalable relational representation.\\


\textit{We thank CSC (IT Center for Science, Finland) for computational resources and the Academy of Finland for the support within the Flagship programme: Finnish Center for Artificial Intelligence (FCAI).}


\begin{footnotesize}

\bibliographystyle{unsrt}
\bibliography{references}

\end{footnotesize}


\end{document}